\documentclass[sigconf, nonacm]{acmart}
\usepackage{float}
\usepackage{xcolor}
\usepackage{subfigure}
\usepackage{bm}
\usepackage{cleveref}
\usepackage{algorithm,algpseudocode}

\AtBeginDocument{%
  }

\acmConference[SC '25]{St. Louis, USA}

\begin{document}

\title[Distributed GNN Explanations]{DistShap: Scalable GNN Explanations \\ with Distributed Shapley Values}

\author{Selahattin Akkas}
\orcid{0000-0001-8121-9300}
\affiliation{
  \institution{Indiana University Bloomington}
  \country{Indiana, USA}
}
\email{sakkas@iu.edu}

\author{Aditya Devarakonda}
\orcid{0000-0002-8251-9150}
\affiliation{
  \institution{Wake Forest University}
  \country{North Carolina, USA}
}
\email{devaraa@wfu.edu}

\author{Ariful Azad}
\orcid{0000-0003-1332-8630}
\affiliation{
  \institution{Texas A\&M University}
  \country{Texas, USA}
}
\email{ariful@tamu.edu}


\begin{abstract}
With the growing adoption of graph neural networks (GNNs), explaining their predictions has become increasingly important. However, attributing predictions to specific edges or features remains computationally expensive.
For example, classifying a node with 100 neighbors using a 3-layer GNN may involve identifying important edges from millions of candidates contributing to the prediction.
To address this challenge, we propose DistShap, a parallel algorithm that distributes Shapley value-based explanations across multiple GPUs. DistShap operates by sampling subgraphs in a distributed setting, executing GNN inference in parallel across GPUs, and solving a distributed least squares problem to compute edge importance scores. DistShap outperforms most existing GNN explanation methods in accuracy and is the first to scale to GNN models with millions of features by using up to 128 GPUs on the NERSC Perlmutter supercomputer.
\end{abstract}



\keywords{GNN explainability, distributed explainable AI, Shapley value}


\maketitle

\section{Introduction}

\begin{figure}[!t]
    \centering
        \includegraphics[width=1.0\linewidth]{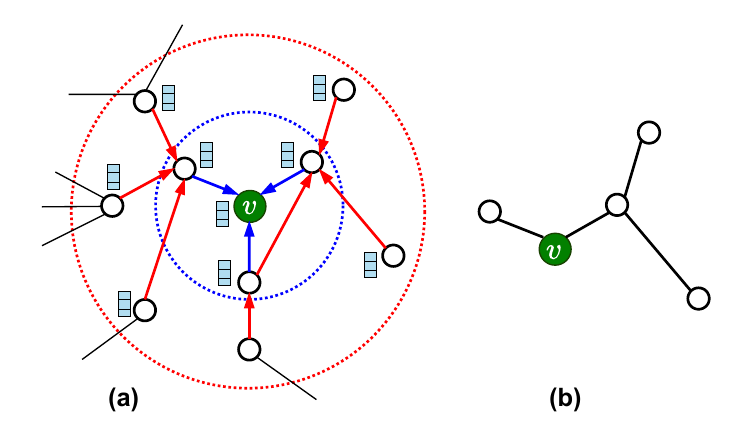}
        \vspace{-15pt}
        \caption{(a) Information propagation in a two-layer GCN: red edges indicate information flow in layer 1, while blue edges represent information flow in layer 2. Other edges do not contribute to the prediction of node $v$. (b) A subgraph $G_s$ containing edges important for $v$'s prediction. An explanation method aims to identify such important edges. \vspace{-5pt}}
        \label{fig:GCN}
\end{figure}

Graph Neural Networks (GNNs) are a class of neural networks specifically designed to handle graph-structured data. 
Their success in tasks such as node classification, link prediction, and recommendation has led to the development of a variety of models, including Graph Convolutional Networks (GCN)~\cite{kipf2016semi_GCN}, GraphSAGE~\cite{hamilton2017inductive_graphsage}, GAT~\cite{velickovic2017graph_gat}, and GIN~\cite{xu2018powerful_gin}. 
These models share a common computational pattern: at each layer, a node in the graph aggregates information from its neighbors and applies a neural network to transform this aggregated information. 
Figure~\ref{fig:GCN} shows an illustration of a two-layer GCN.
Given that a node’s prediction is a complex function of its multi-hop neighbors, it is non-trivial to understand how different nodes, edges, and features contribute to the prediction. GNN explanation methods aim to address this challenge.

{\bf How can predictions of a GNN be explained?}
When explaining a GNN’s prediction, we assume the model has already been trained and focus on understanding how it makes predictions during inference.
For example, in Figure~\ref{fig:GCN}(a), the prediction for $v$ is influenced by all nodes within two hops, while nodes beyond this range have no effect and are excluded.
However, not all neighbors contribute equally to the prediction. Some nodes (and their connecting edges) have a greater impact than others. A GNN explainer seeks to identify which edges are most influential in determining the prediction for $v$. In essence, a basic GNN explanation method returns a subgraph $G_s$, induced by a subset of $v$'s two-hop neighborhood, where the edges in $G_s$ are considered important for the prediction of $v$.
Figure~\ref{fig:GCN}(b) shows such a subgraph.

{\bf How do we measure the effectiveness of explanations?}
Let $G_s$ be the subgraph containing edges important for the prediction of $v$. This explanation is considered effective if removing $G_s$ from the original graph significantly reduces the model’s prediction confidence or accuracy. This measure, known as $Fidelity_{+}$~\cite{2023yuantaxonomy}, captures the idea that edges are important if their removal negatively impacts the model’s ability to make accurate predictions.
A high $Fidelity_{+}$ score indicates an effective explanation method. Conversely, a low $Fidelity_{+}$ score suggests a poor explanation, since removing the identified edges has little to no impact on the model's prediction.

{\bf Challenges in explaining GCNs on large graphs.}
With the growing demand for explainable AI, numerous GNN explanation methods have been proposed in the literature~\cite{ying2019gnnexplainer, 2023yuantaxonomy, akkas2024gnnshap, yuan2020xgnn}. These methods often achieve high $Fidelity_{+}$ scores when applied to nodes with small neighborhoods. However, their effectiveness diminishes as the neighborhood size grows, with many methods struggling to identify truly influential edges.
To illustrate the challenge, consider a node with 1,000 edges in its neighborhood. Its prediction depends on information propagated through these edges. To assess the importance of a single edge, we need to evaluate its marginal contribution -- i.e., how much it changes the prediction when added to each possible subgraph. This approach, grounded in Shapley values from cooperative game theory~\cite{shapley1951notes}, is effective but computationally infeasible, as it requires evaluating $2^{1000}$ subgraphs.
As a result, most existing explanation methods rely on sampling or surrogate models to approximate importance calculations. This introduces a trade-off between explanation accuracy and runtime efficiency, particularly on large graphs.

\begin{figure}[!t]
    \centering
        \includegraphics[width=1.0\linewidth]{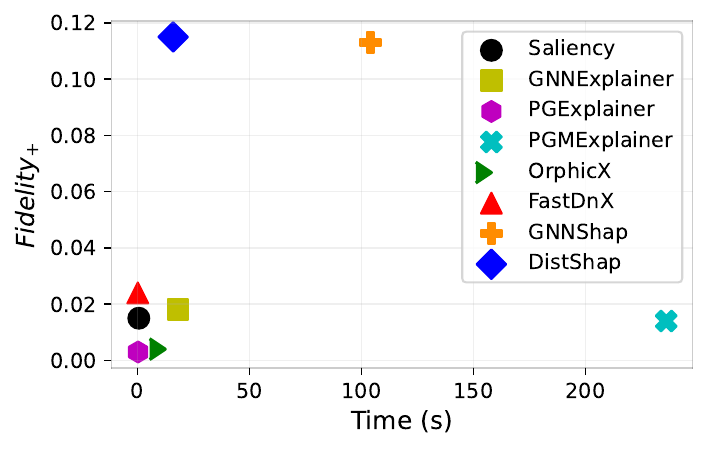}
        \vspace{-15pt}
        \caption{$Fidelity_{+}$ scores for identifying the top 20 most influential edges and corresponding computation times on the Coauthor-CS graph for seven explanation methods. 
        A two-layer GCN was pre-trained before the explanation task. DistShap was run on 8 GPUs, while other methods used a GPU. Higher $Fidelity_{+}$ scores and lower runtimes are desired. \vspace{-5pt} }
        \label{fig:dist-scatter-plot}
\end{figure}

Figure~\ref{fig:dist-scatter-plot} illustrates the trade-off between runtime and explanation quality when identifying the top 20 most influential edges in a paper-authorship graph. 
While most existing methods are fast, they fail to capture truly important edges, as indicated by their low $Fidelity_{+}$ scores. GNNShap~\cite{akkas2024gnnshap} stands out by achieving high $Fidelity_{+}$, but at the cost of significant computation -- requiring evaluation of hundreds of thousands of subgraphs to approximate Shapley values. 
In this paper, we introduce DistShap, a distributed explanation algorithm that accelerates Shapley value computation across multiple GPUs. As shown in Figure~\ref{fig:dist-scatter-plot}, DistShap substantially reduces runtime while maintaining good explanation fidelity.

{\bf Contributions of this paper.}
The main contribution of this paper is the development of DistShap, the first distributed explanation method for GNNs. 
Similar to GNNShap~\cite{akkas2024gnnshap} and GraphSVX~\cite{duval2021graphsvx}, DistShap is based on approximate Shapley value computations. 
However, existing methods are often prohibitively expensive in terms of both computation and memory usage. 
DistShap overcomes these limitations by distributing the explanation workload across hundreds of GPUs. 
This distributed design introduces new challenges -- including scalable subgraph sampling, executing millions of GNN inferences, and solving large-scale least squares problems to compute importance scores. We present efficient parallel algorithms for each of these components, enabling DistShap to scale to 128 NVIDIA A100 GPUs.
We summarize our contributions below:
\begin{itemize}
    \item DistShap is the first distributed algorithm for explaining GNNs. It scales to 128 NVIDIA A100 GPUs and enables explanation of GNNs on graphs that would otherwise be infeasible to explain using other explanation methods.
    \item DistShap achieves higher explanation fidelity than existing state-of-the-art GNN explanation methods.
    \item DistShap includes distributed sampling, batched prediction, and Conjugate Gradients Least-Squares solvers that can be used in other applications.

\end{itemize}

\section{Background \& Related Work}
We consider a graph $G(V, E)$, where $V$ represents the set of nodes and $E$ represents the set of edges. Let $A \in \mathbb{R}^{|V| \times |V|}$ be the sparse adjacency matrix of the graph, where $A_{ij} = 1$ if ${v_i, v_j} \in E$, and $A_{ij} = 0$ otherwise. Additionally, let $X \in \mathbb{R}^{|V| \times d}$ be the matrix of $d$-dimensional node features. Each node is assigned to one of $\mathbb{C}$ classes for node classification tasks. Given a trained GNN model $f$, the predicted class for a node $v$ is $\hat{y} = f(G, X, v)$.

\subsection{Graph Neural Networks}
GNNs utilize a message-passing framework where each layer $l$ involves three key computations~\cite{battaglia2018relational, zhang2020deep, zhou2020graph}. First, messages are propagated between node pairs $(v_i, v_j)$ based on their previous layer representations $h_i^{l-1}$ and $h_j^{l-1}$, as well as the relation $r_{ij}$ between them: $q_{ij}^{l} = \text{MESSAGE}(h_i^{l-1}, h_j^{l-1}, r_{ij})$. Next, these messages are aggregated for each node $v_i$ from its neighbors $\mathcal{N}_{v_i}: Q_{i}^{l} = \text{AGGREGATE}(\{q_{ij}^{l} | v_j \in \mathcal{N}_{v_i}\}) $. Finally, the GNN updates the node representation \( h_i^{l} \) by applying a non-linear transformation to the aggregated message and the node's previous representation: $ h_{i}^{l} = \text{UPDATE}(Q_{i}^{l}, h_i^{l-1})$.

{\bf Computational graph}: 
Once a GNN model $f$ is trained, we can predict the class of a test node $v$ as $\hat{y} = f(G, X, v)$.
However, the prediction does not require access to the entire graph. Specifically, a GNN with  $L$ layers only depends on 
$v$'s 1-hop through $L$-hop neighbors, the edges among them, and any associated node and edge features. 
This $L$-hop subgraph is referred to as the computational graph $G_c(v)$, which contains all the necessary information for predicting $v$.

\subsection{GNN Explanation}
To explain the prediction of node $v$, a GNN explanation method takes a trained GNN model $f$ and the computational graph $G_c(v)$ of node $v$ as inputs.
It then outputs a small subgraph $G_s(v)$ of $G_c(v)$ as the output. 
$G_s(v)$ contains only a few important nodes and edges, thereby providing insight into the reasoning behind the prediction.


\subsection{Shapley Values}
\label{section:background-shapley}
Shapley values~\cite{shapley1951notes} come from cooperative game theory and are used to fairly assign credit to each player (e.g., a feature or graph edge) for their contribution to the outcome of a game (e.g., a model prediction).
Here, we present the general framework for Shapley value computation. Its specific application to GNN explanation will be discussed in the following section.

Let the game be defined by a set of $n$ players $N=\{1,2,...,n\}$ and let $f: 2^N\rightarrow \mathbb{R}$ be a value function that assigns a real number (e.g., model prediction) to each coalition $S \subseteq N$.
The Shapley value $\phi_{i}$ for player $i\in N$ is defined as:
\begin{equation} \phi_{i}= \sum_{S \in N \setminus \{i\}}\frac{|S|!(n - |S| -1)!}{n!}\left [f(S\cup {i}) - f(S) \right ], \label{eq:shapley-exact} \end{equation}
where $f(S\cup {i}) - f(S)$ is the marginal contribution of player $i$ to the coalition $S$.
Therefore, Eq.~\ref{eq:shapley-exact} computes the weighted average of $i$'s marginal contributions across all possible coalitions that exclude $i$.
For GNN explanations, edges in the computational graph are considered players, and each subgraph represents a coalition of these players. 
Therefore, a computational graph with $n$ edges can generate $2^n$ possible subgraphs or coalitions. 

Given Shapley values for all players, the model’s prediction for the full set of players $f(N)$ can be decomposed as:
\begin{equation}
f(N) = \phi_0 + \sum_{i\in N} \phi_i, \label{eq:exact-model}
\end{equation}
where $\phi_0$ is the expected prediction with no players. 

Since evaluating $2^n$ coalitions in Eq.~\ref{eq:shapley-exact} is computationally infeasible, practical approaches rely on sampling a subset of coalitions to approximate the contribution of each player.
Kernel SHAP~\cite{lundberg2017shap} approximates Shapley values by sampling $k$ coalitions ($k \ll 2^{n}$) and training a linear surrogate model $g$.
We can solve this linear model 
by solving the following weighted linear regression:

\begin{equation}
\min_{\phi} \sum_{S\subseteq N} w(S)\left [f(S) - (g(S))\right]^{2}, 
\label{eq:kernelshap}
\end{equation}
where the kernel weight is defined as follows
\begin{equation}
w(S)= \frac{n-1}{\binom{n}{|S|}|S|(n-|S|)}.
\end{equation}
In this setting, the kernel weight assigns individual coalition weights based on coalition size, giving more weight to smaller and larger coalitions due to the ease of observing individual effects. 

Eq.~\ref{eq:kernelshap} needs to sample a set of $k$ coalitions from all possible $2^n$ coalitions. 
Let $M$ be a $k\times n$ binary mask matrix, where $k$ is the number of sampled coalitions and $n$ is the number of players.
The $i$th row of $M$  specifies the players for the $i$th coalition, with  $M[i,j]=1$ indicating that the $j$th player is included in the $i$th coalition.
Let $W$ be a diagonal matrix storing weights of all samples along the diagonal, and $\hat{y}\in \mathbb{R}^k$  the model prediction for this coalition. 
Then, solving Eq.~\ref{eq:kernelshap} is equivalent to 
solving the weighted least squares problem: $\phi = (M^{T}WM)^{-1}M^{T}W\hat{y}$.

The remaining question is how to sample coalitions from $2^n$ coalitions. 
The Kernel SHAP~\cite{lundberg2017shap} approach distributes samples based on the assigned weights for coalitions of a given size. Let $\rho_{|S|}$ be the total weight of all coalitions of size $|S|$. Then, the number of samples $k_{|S|}$ of size $|S|$ is computed as follows:

\begin{equation}
\rho_{|S|}= \frac{n-1}{|S|(n-|S|)},\ \ \ \ \ \ 
k_{|S|}= k * \frac{\rho_{|S|}}{\sum_{i=1}^{n-1}\rho_{i}}
\label{eq:shap-samp-dist}
\end{equation}
DistShap adopts this non-uniform sampling strategy, which is further detailed in the following section.






\subsection{Related Work}
Most GNN explanation methods~\cite{2023yuantaxonomy} explain how a GNN makes a specific prediction for a given input (e.g., explaining a node's classification). 
These methods can be broadly grouped into several categories based on the explanation generation mechanism.
Gradient-based methods determine importance using input gradients, activation patterns, or attention weights.
Notable works include applying Saliency \cite{pope2019explainability, baldassarre2019explainability}, Guided Backpropagation \cite{baldassarre2019explainability}, Class Activation Mapping (CAM) \cite{pope2019explainability}, GradCAM \cite{pope2019explainability}, and Integrated Gradients \cite{sundararajan2017integratedgrads} to the graph domain. 
While these methods are typically very fast, they often perform poorly at identifying important edges or features in GNNs~\cite{shrikumar2017deeplift}.


Perturbation-based methods evaluate the importance of nodes, edges, or subgraphs by systematically modifying or removing them and measuring the resulting change in the model's prediction. Methods in this category include GNNExplainer \cite{ying2019gnnexplainer}, PGExplainer \cite{luo2020pgexplainer}, GraphMask \cite{schlichtkrull2021graphmask}, SubgraphX \cite{yuan2021subgraphx}, EdgeSHAPer \cite{mastropietro2022edgeshaper}, GraphSHAP \cite{perotti2023graphshap}, GStarX \cite{zhang2022gstarx}, FlowX \cite{gui2024flowx}, and Zorro \cite{funke2022zorro}. These methods are often model-agnostic (treating the GNN as a black box) but can be computationally intensive, especially when exploring numerous perturbations or complex combinatorial spaces.

Surrogate methods train a simpler, inherently interpretable surrogate model (e.g., a linear model) to mimic the GNN's behavior around a specific prediction. Examples include Graph-Lime \cite{huang2022graphlime}, RelEx \cite{zhang2021relex}, GraphSVX \cite{duval2021graphsvx}, PGM-Explainer \cite{vu2020pgmexplainer}, FastDnX \cite{pereira2023fastdnx}, GNNShap~\cite{akkas2024gnnshap}. The fidelity of the explanation depends on how well the surrogate model approximates the original GNN.

\begin{figure*}[!t]
  \centering
    \includegraphics[width=\textwidth]{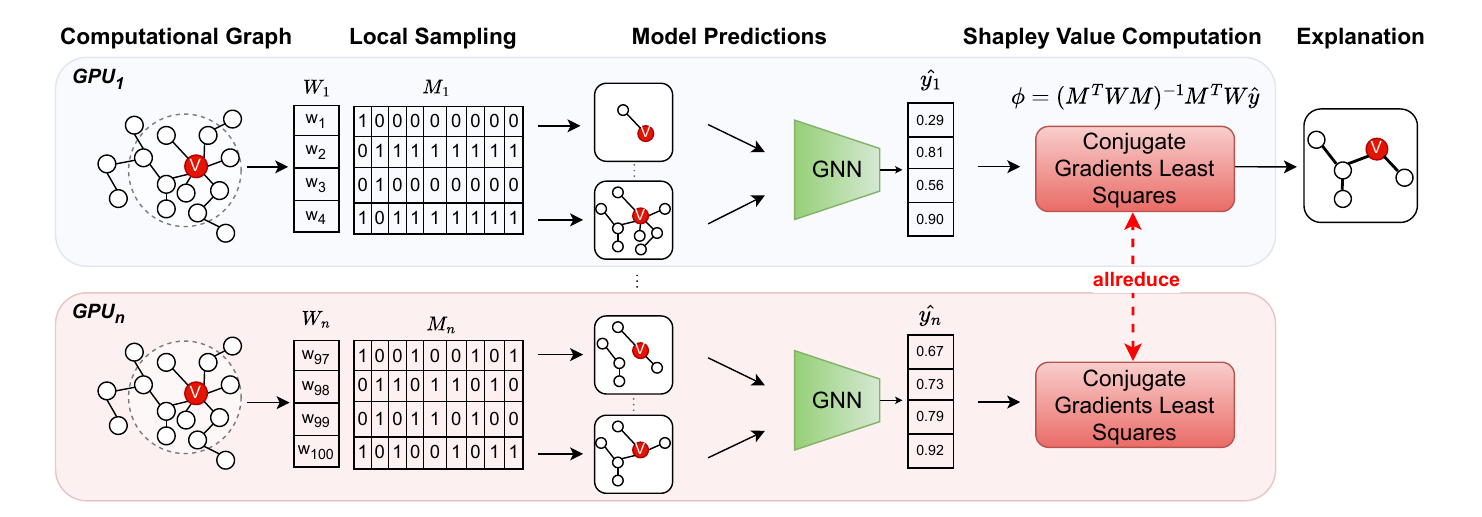}
    \vspace{-15pt}
    \caption{DistShap computes Shapley values for all edges in the computational graph of a target node $v$, using a two-layer GCN model as an example. First, the computational graph of $v$ is replicated across multiple GPUs. Each GPU independently generates a subset of samples using its local mask matrix and performs model predictions for $v$ on the corresponding sampled subgraphs. Once all predictions are completed, the GPUs collaboratively compute approximate Shapley values for every edge in the computational graph. Finally, DistShap identifies and returns the top-k most important edges based on Shapley values.}
  \label{fig:distshapmain}
\end{figure*}

{\bf Game-Theoretic methods.}
GRAPHSHAP \cite{perotti2023graphshap} and EdgeSHAPer \cite{mastropietro2022edgeshaper} focus on graph classification by assigning importance scores to edges or graph motifs. 
GraphSVX~\cite{duval2021graphsvx} explains node classification by using an approximation strategy that heavily samples small and large coalitions, potentially leading to estimations of suboptimal explanations. 
Similar to GraphSVX, GNNShap~\cite{akkas2024gnnshap} estimates Shapley values for edges by sampling coalitions. Although GNNShap produces high-fidelity explanations, it is computationally expensive.
SubgraphX \cite{yuan2021subgraphx} aims to identify the most explanatory subgraph using Shapley values, employing a Monte Carlo Tree Search (MCTS) to explore possible subgraphs. 
GStarX \cite{zhang2022gstarx} also uses MCTS but utilizes Hamiache-Navarro \cite{hamiache2020hnvalue} values.

{\bf Distributed Shapley Value Computation.}
There have been several distributed Shapley value computation attempts. For instance, \cite{Ebrahimi2022ScalingSHAP} and \cite{Marsh2020Shparkley} utilize Apache Spark to speed up Shapley value computations. However, these methods are limited to a relatively small number of players (i.e., up to 50). GPUTreeShap \cite{mitchell2022gputreeshap} utilizes multi-GPU for ML explanations, yet it can only explain tree-based models. While the attempts are beneficial, our work addresses the challenges of solving GNN explanations with thousands of players.

\section{GNN Explanations with Distributed Shapley}
\subsection{Overview of DistShap}
A Shapley-based GNN explanation typically involves five steps: {\bf (1) Define the game}. For GNN explanations, players are edges in the computational graph $G_c(v)$, coalitions are subgraphs of $G_c(v)$, and the value function is the output of the trained GNN model $f$. {\bf (2) Sample coalitions.} For GNN explanation, each sampled coalition is a subgraph of $G_c(v)$. 
{\bf (3) Model prediction.} For each sampled subgraph or coalition, we compute the output of the GNN model.
{\bf (4) Estimate Shapley values.} We then estimate Shapley values by solving the weighted least squares problem.
{\bf (5) Rank and select important edges.} The final Shapley values indicate the importance of each edge of the computational graph.
We select the top-k important edges to form the explanation subgraph.

DistShap parallelizes the sampling, model prediction, and Shapley value computation across multiple GPUs. 
Figure~\ref{fig:distshapmain} shows an overview of DistShap for the task of explaining the prediction of node $v$.
We describe each step in the 
following sections.

\begin{figure}[!t]
    \centering
        \includegraphics[width=1.0\linewidth]{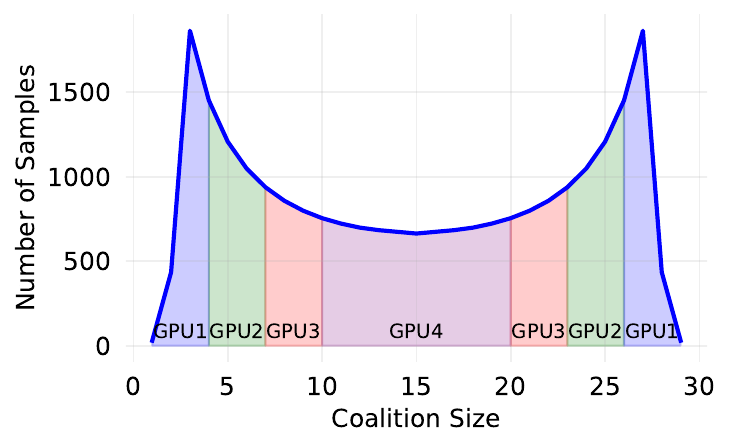}
        \vspace{-15pt}
        \caption{DistShap distribution of 25,000 samples among 4 GPUs. Here, the computational graph has 30 edges, and hence the coalition size ranges from 1 to 29. \vspace{-5pt}}
        \label{fig:abl-gpusampledist}
\end{figure}





\subsection{Distributed Subgraph Sampler}
\subsubsection{Replicating the computational graph}
The first step of DistShap is to extract the computational graph $G_c(v)$ of the target node $v$ and replicate it on each GPU. 
Although the entire graph can be extremely large and may not fit into GPU memory, the computational graphs of individual nodes are relatively small and can easily fit within GPU memory. 
Our choice of replicating the computational graph ensures that sampling and model prediction can run independently on each GPU without inter-GPU communication. 
This design enables highly efficient sampling and allows DistShap to scale nearly linearly with the number of GPUs. 

Since we explain only a small subset of nodes (50 in our experiments), generating their computational graphs is not computationally intensive. 
When the full graph does not fit in GPU memory, we extract the computational graphs on the CPU. 

\subsubsection{Distributed Subgraph Sampling}
In the sampling step, a subgraph (i.e., a coalition) is randomly selected from the computational graph. 
In this step, two decisions must be made: (1) how to generate 
$k$ representative samples from the exponentially large space of $2^n$ subgraphs, and (2) how to efficiently distribute
$k$ samples across 
$p$ GPUs in a computationally efficient and load-balanced manner.

{\bf Non-uniform Sampling.}
Prior work~\cite{akkas2024gnnshap, duval2021graphsvx} has shown that uniform sampling from the $2^n$ possible subgraphs does not provide good explanations for large computational graphs. 
Hence, we follow previous approaches and adopt a non-uniform sampling strategy guided by Eq.~\ref{eq:shap-samp-dist} and  Fig.~\ref{fig:abl-gpusampledist}.
This distribution favors sampling coalitions with very few or very many edges, as these tend to provide more informative signals for explaining the prediction. 
However, mid-sized coalitions are not ignored, ensuring a balanced coverage across the coalition space, as shown in Fig.~\ref{fig:abl-gpusampledist}.
Moreover, the sampling strategy is symmetric: for every sampled subgraph with $a$ edges, its complementary subgraph with $n-a$ edges is also included. This pairing of complementary samples helps the algorithm converge more quickly to the true Shapley values.

{\bf Partitioning samples across GPUs.}
When generating samples, each GPU should generate $k/p$ samples, where $k$ is the total number of samples, and $p$ is the number of GPUs. 
A naive approach that assigns samples sequentially -- letting the $i$th GPU handle samples in the range $i\cdot k/p:(i+1)k/p$ -- can lead to load imbalance in subsequent steps. 
For example, the leftmost samples in Fig.~\ref{fig:abl-gpusampledist} have fewer edges, which results in lower memory usage and faster model predictions.
To address this imbalance, we leverage the symmetry of the sampling distribution. In our approach, each GPU generates both members of a complementary sample pair, as illustrated in Fig.~\ref{fig:abl-gpusampledist}. 
Since each pair of complementary subgraphs contains a total of $n$ edges, this strategy ensures that each GPU processes exactly $k/p$ samples and a total of $\frac{nk}{2p}$ edges.  
This results in a perfectly balanced workload across all GPUs.

\begin{figure}[!t]
    \centering
        \includegraphics[width=1.0\linewidth]{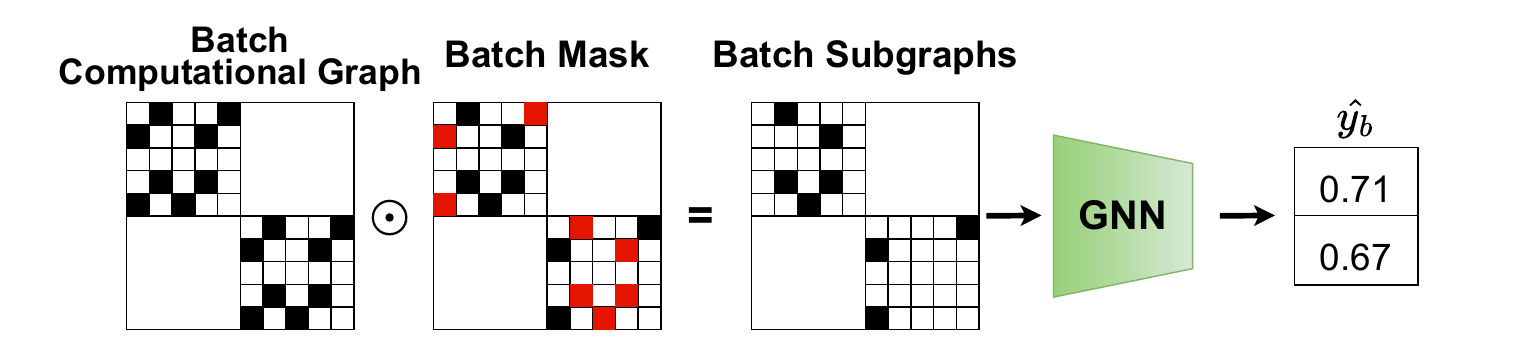}
        \vspace{-5pt}
        \caption{Batch masking and model prediction by stacking the computational graph diagonally. Red cells show edges that are not in the coalition. \vspace{-5pt} }
        \label{fig:batching}
\end{figure}

{\bf Sampling via a mask matrix.}
Instead of directly sampling subgraphs from the computational graph, we use a $k\times n$ binary mask matrix $M$ mask matrix to generate samples.
The $i$th row of $M$  specifies the edge configuration for the $i$th sample, with  $M[i,j]=1$ indicating that the $j$th edge is included in the $i$th sample.
Given the paired nature of the sampling, $M$ is 50\% sparse. 
Based on our partitioning strategy shown in Fig.~\ref{fig:abl-gpusampledist}, each GPU generates $k/p$ rows of $M$.
Generating the local submatrix of $M$ is equivalent to generating random numbers from a Bernoulli distribution. 
For example, if two randomly selected columns in the $i$th row are set to 1, that row represents a sampled subgraph containing the edges corresponding to the non-zero columns. 
Since this step involves generating large numbers of random values, it is embarrassingly parallel and can be efficiently executed across GPU threads.
In practice, we generate only $k/2$ subgraphs, as the complementary subgraph for each can be derived deterministically without generating additional random numbers. 
Once the mask matrix is created, we generate subgraphs by simply applying a binary mask over the edge list of the computational graph.
This two-step approach (illustrated in Fig.~\ref{fig:distshapmain} and Fig.~\ref{fig:batching}) is highly amenable to parallel execution and enables lightning-fast sampling, as demonstrated in the results section.

\subsubsection{Time and Memory Complexity}
The memory required to store the submatrix of $M$ and corresponding subgraphs is $\mathcal{O}(n\cdot k/p)$ per GPU.
As $t$ threads in a GPU can generate samples independently, the time complexity is $\mathcal{O}(\frac{n\cdot k}{t\cdot p})$.

\subsection{Distributed Model Prediction}
We perform GNN model prediction for each sampled subgraph.
For instance, in an explanation task with 600K samples distributed across 10 GPUs, each GPU is responsible for 60K predictions.
Executing these predictions sequentially is inefficient, as each individual prediction may not fully utilize the GPU's computational capacity.
We address this issue by batching multiple subgraphs together and performing inference on the entire batch simultaneously.

Fig.~\ref{fig:batching} illustrates the batching process with a batch size of 2.
First, we construct a block-diagonal matrix by placing two copies of the computational graph's adjacency matrix along the diagonal.
Next, we create a corresponding block-diagonal mask matrix, where each block contains the edge mask for one sample.
An element-wise product of these two matrices yields another block-diagonal matrix representing the sampled subgraphs.
Finally, the GNN model performs inference on this entire batch of subgraphs simultaneously.
Batching improves parallel efficiency but comes at the cost of increased memory usage.

{\bf Time and memory complexity.} Suppose we are using a two-layer GCN with a hidden dimension of $d$.
Let $b$ be the number of batches, and $|V_c|$ be the number of nodes in the computational graph.
Then, the time for batched model prediction on each GPU is

\begin{equation}
    O\left( b\cdot (n \cdot d + |V_c| \cdot d^2 ) \right).
\end{equation}


The memory requirement for a batch is
\begin{equation}
    O\left( b\cdot (n +  |V_c| + |V_c| \cdot d \right)).
\end{equation}

Increasing the batch size leads to a linear increase in GPU memory usage.
However, it significantly improves GPU utilization, thereby reducing overall computation time -- up to the point where resource limits are reached.
Our experiments (Fig.~\ref{fig:dist-ablationbatchsize}) indicate that a batch size of 50 offers an optimal trade-off, delivering strong runtime performance while keeping memory usage manageable.

\begin{algorithm}
\caption{Distributed Conjugate Gradients Least-Squares}
\label{alg:distcgls}
\begin{algorithmic}[1]
    \State \textbf{Input:} $M_p \in \mathbb{R}^{k/p \times n},W_p \in \mathbb{R}^{k/p \times k/p},\hat{y_p} \in \mathbb{R}^{k/p},x_0 \in \mathbb{R}^n$
     \Comment{Subscript $p$ denotes GPU-local inputs}
    \State $M_p = \sqrt{W_p}M_p$
    \State $\hat{y_p} = \sqrt{W_p}\hat{y_p}$
    \State $r_0 = \hat{y_p}$
    \State $s_0 = M_p^\intercal \hat{y}$
    \State $s_0 = \Call{All\_reduce}{s_0}$
    \Comment{$n$-dimensional vector reduction}
    \State $u_0 = s_0$
    \For{$i = 1, 2 \ldots$ until convergence}
        \State{$v_k = M_p u_{i-1}$}
        \State{$\delta = \Vert v_i \Vert_2^2$}
        \State{$\delta = \Call{All\_reduce}{\delta}$}
        \Comment{Scalar reduction}
        \State{$\theta = \frac{\Vert s_{i-1}\Vert_2^2}{\delta}$}
        \State{$x_i = x_{i-1} + \theta u_{i-1}$}
        \State{$r_i = r_{i-1} - \theta v_{i-1}$}
        \State{$s_i = M_p^\intercal r_i$}
        \State{$s_i = \Call{All\_reduce}{s_i}$}
        \Comment{$n$-dimensional vector reduction}
        \State{$p_i = s_i + \frac{\Vert s_{i-1}\Vert_2^2}{\Vert s_i \Vert_2^2} u_{i-1}$}
    \EndFor \\
    \Return{ $x_i$}
\end{algorithmic}
\end{algorithm}

\subsection{Distributed Shapley Computations}
Shapley values are the solution to the following weighted least-squares problem:
$\phi = (M^{T}WM)^{-1}M^{T}W\hat{y}$, where $M$ is the mask matrix, $W$ is a diagonal weights matrix, and $\hat{y}$ is the vector of model predictions.
Since $k \gg n$, a parallel direct method to obtain $\phi$ is to store $M$ in a 1D-row layout and perform the following computations in parallel: compute $G = M^{T}WM$, $r = M^{T}W\hat{y}$ and solve the linear system $\phi = G^{-1}r$ sequentially.
Communication is required to sum-reduce $G$ across all GPUs, which requires communicating $n^2$ entries.
However, for large enough $n$, solving the linear system sequentially and communicating all of $G$  is likely to become a bottleneck.
Other direct methods based on LU, QR, or their tall-skinny variants (e.g., TSQR~\cite{demmel2012communication}) require complex 2D block cyclic data layouts (e.g., LU and QR) and also require parallel triangular solve routines (LU, QR, and TSQR), which are challenging to parallelize and scale.

Instead, we utilize a distributed-memory implementation of the Conjugate Gradients Least Squares (CGLS) method \cite{hestenes1952methods,cgls,cgls-alg} to solve the weighted least-squares problem iteratively.
CGLS is a Krylov method that adapts CG to the least squares problem by implicitly solving the normal equations, and is more numerically stable than applying a parallel CG algorithm to solve $G^{-1}r$.
Each iteration of CGLS requires a sequence of matrix-vector, vector, and scalar operations that are simpler to parallelize.
While Krylov methods are often applied to solve sparse linear systems, we apply CGLS to solve a dense least-squares problem.
Since $M$ has a sparsity of 50\%, we store it in a dense format, which enables us to use the more efficient dense matrix-vector multiplication kernel on the GPU instead of an SpMV.
We parallelize CGLS by storing $M$ in a 1D row layout such that each GPU stores $k/p$ rows of $M$ locally.
We partition $\hat{y}$ similarly but replicate all $n$-dimensional vectors used in CGLS.
Under a 1D row layout, norm computations of $k$-dimensional vectors and matrix-vector products with $M^\intercal$ require reductions across GPUs to ensure sequential consistency, as each GPU stores $k/p$ fraction of matrix/vector entries. We show the distributed CGLS algorithm in \Cref{alg:distcgls}. 

Each iteration of the distributed CGLS algorithm performs orthogonalization of the solution ($x_i$) and residual ($r_i$) vectors associated with the left and right subspaces of $M$.
Scalar quantities (e.g. $\theta$) represent step-sizes use to update the conjugate directions $s_i$ and $p_i$.
Without round-off error, CGLS requires $n$ iterations to find the solution $\phi$.
However, in practice, the number of iterations depends on the condition number of the input matrix, $M$.
Since \Cref{alg:distcgls} is iterative, communication is required every iteration to compute the scalar quantity, $\delta$, and the $n$-dimensional vector quantity, $s_i$.
All remaining operations are local to each GPU and do not require communication.
The communication complexity associated with \Cref{alg:distcgls} using Hockney's model \cite{hockney} is
\begin{equation*}
    T_{comm} = \alpha \cdot \mathcal{O}(i \cdot \log{p}) + \beta \cdot \mathcal{O}(i\cdot n),
\end{equation*}
where $i$ is the number of iterations to find the solution, $\alpha$ and $\beta$ are hardware latency and inverse bandwidth parameters, respectively.
We show that \Cref{alg:distcgls} is faster than solving the weighted least-squares problem using a parallel direct method despite the additional latency cost.
Note that in theory, $i = n$ is sufficient for CGLS to converge, so the total bandwidth cost is equivalent to communicating $G$ in the previously described direct method.
We compare the performance of \Cref{alg:distcgls} against a parallel direct approach in \Cref{sec:ablation-studies}.

\section{Experiments}

\subsection{Datasets}
We use six real-world datasets for the experiments: Coauthor-CS and Coauthor-Physics \cite{dataset_coauthor}, DBLP \cite{dataset_citationfull}, Reddit~\cite{hamilton2017inductive_graphsage},  ogbn-arxiv, and ogbn-products~\cite{hu2020ogbn}. Coauthor-CS and Coauthor-Physics are co-author networks where the nodes represent authors, the edges indicate coauthorships, and the node attributes are keywords from the papers. DBLP is a citation network where nodes are papers, edges are citations, and node features are bag-of-words embeddings. ogbn-arxiv is another citation network whose embeddings are obtained by the skip-gram model \cite{mikolov2013distributed}. 
ogbn-products is an Amazon co-purchase network where nodes represent products and edges represent co-purchased products. Node features are bag-of-words features followed by dimension reduction. 
In the Reddit dataset, each node represents a post from a subreddit. Node features include text-based attributes, while the labels correspond to the subreddits the posts belong to.
Table \ref{table:dataset} summarizes the datasets.


\begin{table}[!tb]
\centering
\caption{Dataset Summary}
\vspace{-10px}
\label{table:dataset}
\resizebox{\linewidth}{!}{
\begin{tabular}{lcccc}
\hline
\textbf{Dataset} & \textbf{Nodes} & \textbf{Edges} & \textbf{Features} & \textbf{Classes} \\
\hline
Coauthor-CS      & 18,333         & 163,788        & 6,805             & 15               \\
Coauthor-Physics & 34,493         & 495,924        & 8,415             & 5                \\
DBLP             & 17,716         & 105,734        & 500               & 4                \\
ogbn-arxiv       & 169,343        & 2,315,598      & 128               & 40               \\
ogbn-products    & 2,449,029      & 123,718,280    & 100               & 47               \\
Reddit           & 232,965        & 114,615,892    & 602               & 41         \\
\hline
\end{tabular}
}
\end{table}

\subsection{GNN Models}
Since all GNN explanation models are designed to explain a pre-trained model, they can be applied to any GNN architecture.
In our experiments, we use a two-layer GCN  and train it for the node classification task.
We use 128 hidden dimensions for the Reddit, ogbn-products, and ogbn-arxiv datasets and 64 hidden dimensions for the Coauthor and DBLP datasets.
We use the ReLU activation and optimize the model using cross-entropy loss.
We also show results with a two-layer GAT model.

\subsection{Explanation Setting}
In our experiments, we select 
50 nodes from the test data. For Reddit, ogbn-arxiv, and ogbn-products, we select nodes with 50,000 to 100,000 edges in their computational graph (i.e., the number of players is at least 50K). 
For the Coauthor and DBLP datasets, we select nodes with at least 1000 edges in their computational graphs.

\subsection{Evaluation Metrics}

$Fidelity_{+}$ and $Fidelity_{-}$ are two widely used metrics \cite{2023yuantaxonomy} to value explanation methods. 
Let $G_c \subseteq G$ be the computational graph for node $v$ and $G_s \subseteq G_c$ be the subgraph identified by the explainer as important for predicting node $v$.
Then, Eq.~\ref{eq:fid-pos} shows how $Fidelity_{+}$ is calculated for node $v$:
\begin{equation}
Fidelity_{+}(v) = \left |f(G_c) - f(G_c\setminus G_s) \right |
\label{eq:fid-pos}
\end{equation}
Here, $f(G_c)$ is the prediction score (associated with the class predicted by the model) for $v$ on the original graph, and $f(G_c\setminus G_s)$ is the prediction score for the same node when the important edges in $G_s$ are removed from the graph.
Eq.~\ref{eq:fid-pos} shows how $Fidelity_{-}$ is calculated for node $v$:
\begin{equation}
Fidelity_{-}(v) = \left |f(G_c) - f(G_s) \right |
\label{eq:fid-neg}
\end{equation}



Based on these equations, $Fidelity_{+}$ measures how model prediction changes when important edges are removed, while $Fidelity_{-}$ measures how model prediction changes when least important edges are removed. 
A good explanation should yield high  $Fidelity_{+}$ scores, indicating the importance of selected edges,   and low $Fidelity_{-}$ scores, reflecting the irrelevance of unselected edges.


\begin{figure*}[!t]
    \centering
        \includegraphics[width=1.0\textwidth]{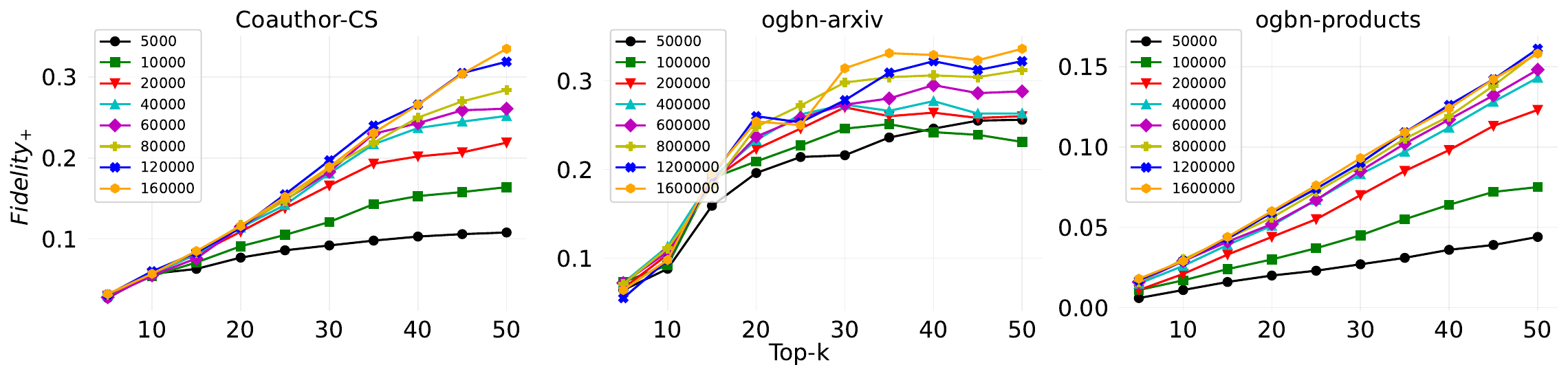}
        \caption{Average $Fidelity_{+}$ scores (higher is better) for the top-k most important edges identified by DistShap across various sample sizes. Increasing the number of samples generally improves the results, up to a point of saturation. For smaller graphs like Coauthor-CS, we used up to 160K samples, while for larger graphs, we used up to 1.6 million samples.}
        \label{fig:dist-nsamples}
\end{figure*}

\subsection{GNN Explanation Baselines}

\begin{itemize}
    \item Saliency \cite{pope2019explainability, baldassarre2019explainability}: computes output gradients with respect to input features and sums them to estimate node importance.
    \item GNNExplainer~\cite{ying2019gnnexplainer}: leverages mutual information to identify significant edges and features within the graph.
    \item PGExplainer~\cite{luo2020pgexplainer}: PGExplainer employs mutual information and trains a neural network to generate explanations. 
    \item PGMExplainer~\cite{vu2020pgmexplainer}: uses a probabilistic graphical model to determine the importance of nodes. 
    \item OrphicX~\cite{lin2022orphicx}: produces causal explanations using a generative surrogate model. We train OrphicX for 50 epochs.
    \item FastDnX \cite{pereira2023fastdnx}: uses decomposition of linear surrogate model to explain GNN models. We train FastDnX for 1,000 epochs.
\end{itemize}

\subsection{Test Environment}
We run all experiments on the Perlmutter supercomputer at NERSC. Each node has two AMD EPYC 7713 64-Core Processors, 512 GB main memory, and four Nvidia A100 80GB GPUs. We use Python 3.12.3, Cuda 12.4, PyTorch 2.4.1, and PyTorch Geometric 2.6.0.

\section{Results}

\subsection{Impact of Sampling on Explanation Fidelity}
At first, we analyze the effect of the number of samples on DistShap's ability to find important edges. 
Fig. \ref{fig:dist-nsamples} shows the average $Fidelity_{+}$ score achieved by DistShap for a small-scale (Coauthor-CS) and two medium- to large-scale datasets  (ogbn-arxiv and ogbn-products). 
The $Fidelity_{+}$ score generally increases for a fixed sample size as more important edges are included in the explanation. 
However, in most cases, selecting 30 to 50 of the most important edges is sufficient to explain the prediction effectively. 
Regarding sample size, larger samples tend to improve $Fidelity_{+}$ up to a saturation point. 
For example, in Coauthor-CS, the $Fidelity_{+}$ score shows little improvement beyond 60K samples, while for ogbn-arxiv and ogbn-products, it plateaus beyond 600K samples. 
Since increasing the number of samples raises the computational and memory cost, we use 600K and 60K samples for large and smaller graphs, respectively.



\begin{figure*}[!t]
    \centering
        \includegraphics[width=0.95\linewidth]{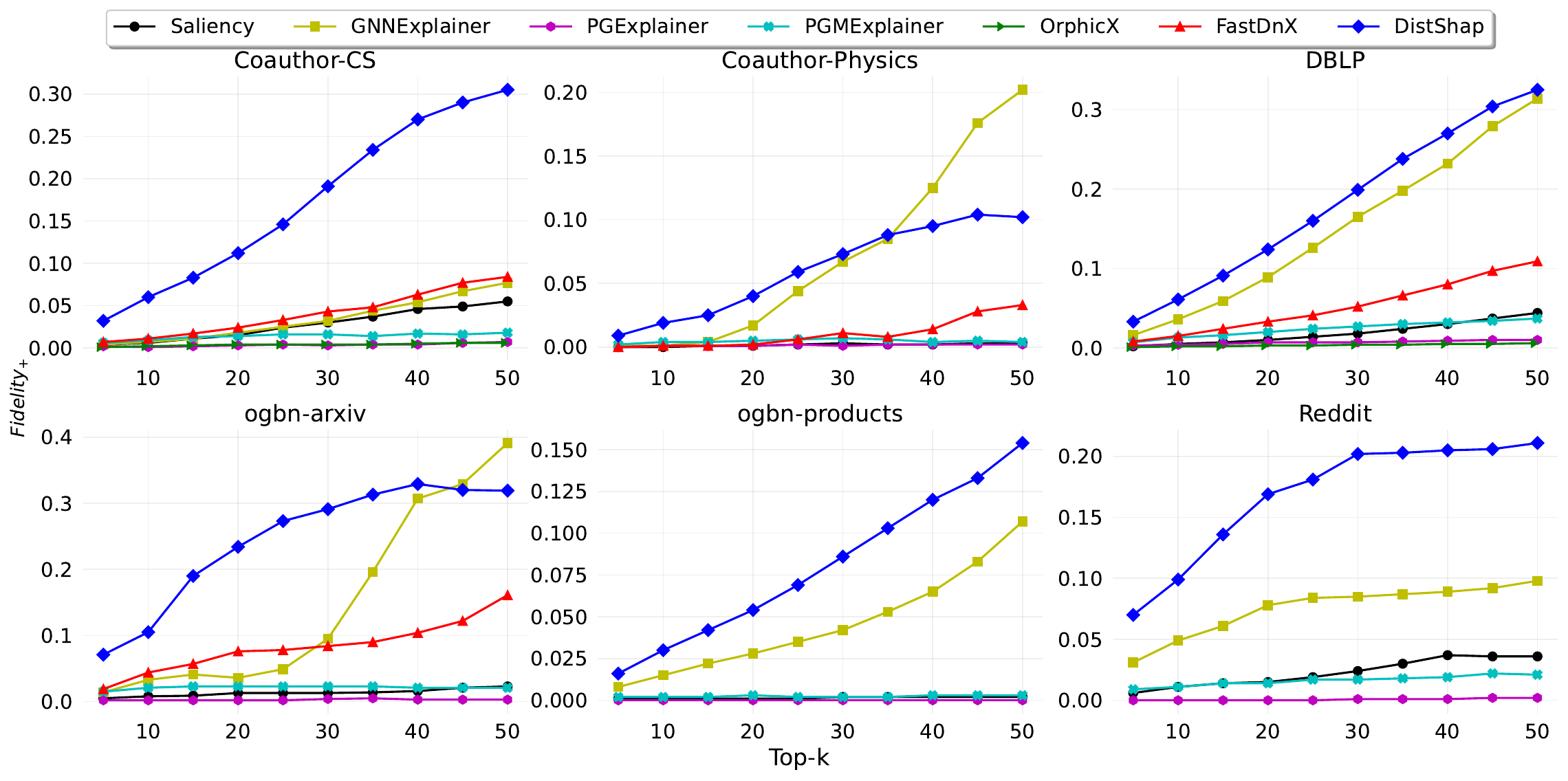}
        \vspace{-10pt}
        \caption{Average $Fidelity_{+}$ scores for top-k most important edges (higher is better) with a two-layer GCN model. DistShap gives the best or near-best results for all values of $k$.}
        \label{fig:dist-fidpos}
\end{figure*}
\begin{figure*}[!t]
    \centering
        \includegraphics[width=.95\linewidth]{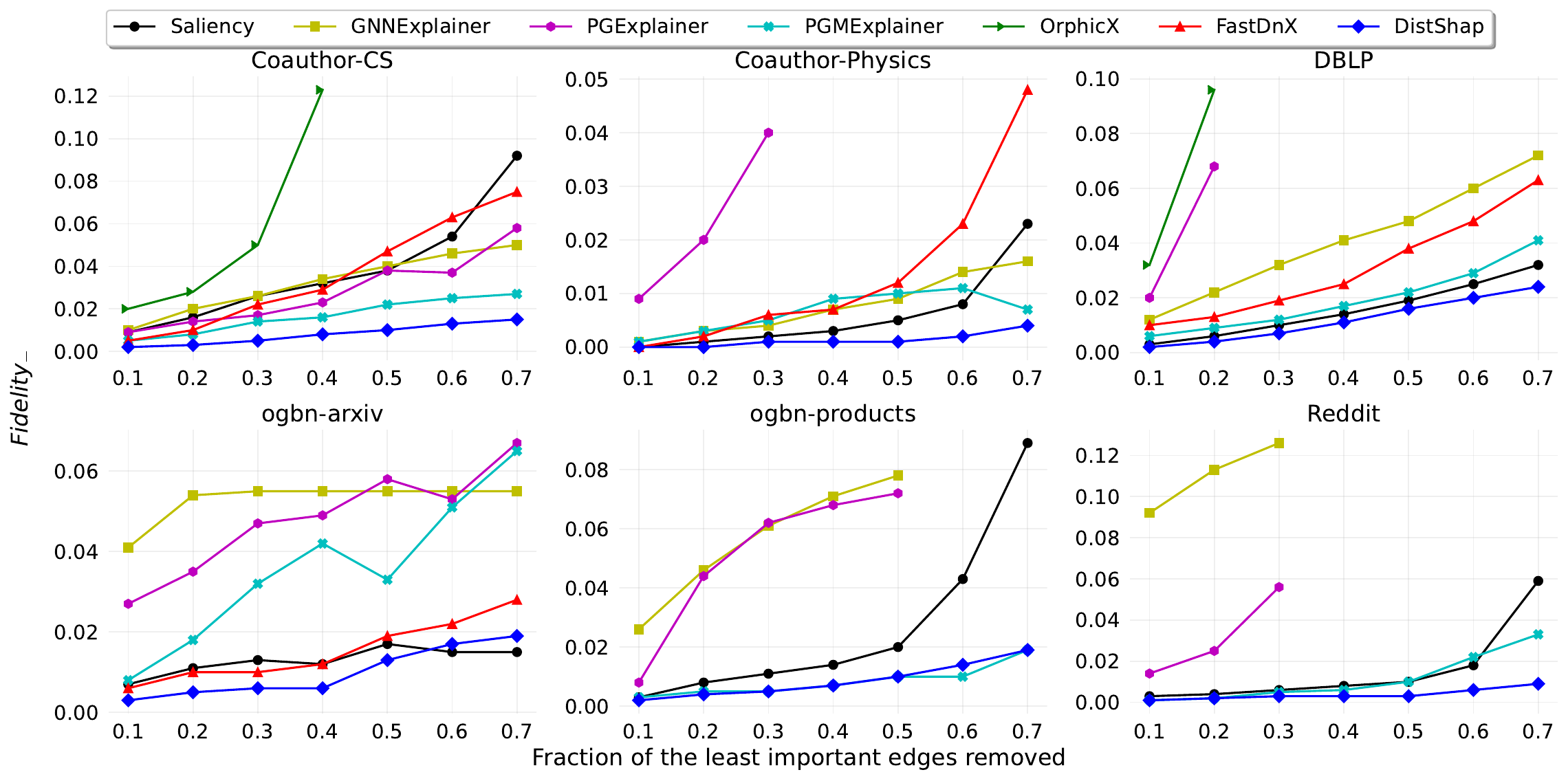}
        \vspace{-10pt}
        \caption{Average $Fidelity_{-}$ scores for various sparsity levels (lower is better) with a two-layer GCN model. 
        A 0.7 sparsity means removing 70\% of the least important edges. DistShap gives the best or near-best results for all sparsity levels.}
        \label{fig:dist-fidneg}
\end{figure*}

\subsection{Fidelity Comparisons with Other Baselines}
Fig.~\ref{fig:dist-fidpos} presents the average $Fidelity_{+}$ scores for 50 test nodes with a two-layer GCN model. 
In this figure, “top-30” refers to the effectiveness of the top 30 edges identified by each explanation method, with higher scores indicating better identification of important edges. We observe that DistShap consistently outperforms baseline methods across most settings by accurately identifying influential edges. GNNExplainer shows slightly better performance only beyond the top-40 edges on ogbn-arxiv and after the top-30 edges on Coauthor-CS. However, its overall performance is less reliable, especially in terms of $Fidelity_{-}$ scores discussed next.

Fig.~\ref{fig:dist-fidneg} presents the average $Fidelity_{-}$ scores for 50 test nodes with a two-layer GCN model. 
In this figure, a sparsity level of 0.7 means that 70\% of the least important edges are removed from the computational graph. 
We observe that DistShap gives the best or near-best $Fidelity_{-}$ scores for all sparsity levels by accurately identifying unimportant edges.
While PGMExplainer exhibits competitive $Fidelity_{-}$ scores on Reddit and ogbn-products, its performance is not robust for other datasets.

We note that other Shapley-based explanation methods, such as GNNShap and GraphSVX, may produce results comparable to DistShap. 
However, these methods are either too slow or run out of memory when applied to large-scale graphs. 

\begin{table*}[]
\caption{Total explanation times for 50 nodes. Training times of surrogate models are provided in parentheses. DistShap uses 8 GPUs for Coauthor and DBLP and 128 GPUs for other datasets. (Err: CUDA error - insufficient resources; OOM: Out Of Memory)
}
\label{table:dist-explanation-times}
\resizebox{\textwidth}{!}{
\begin{tabular}{lcccccc}
\hline
             & Coauthor-CS & Coauthor-Physics & DBLP   & ogbn-arxiv & ogbn-products & Reddit   \\
\hline
Saliency     & 0.56        & 0.76             & 0.47   & 0.98       & 0.87          & 1.37     \\
GNNExplainer & 18.05       & 18.59            & 15.97  & 61.30      & 79.01         & 1,092.40 \\
PGExplainer  & 0.22 (392.49)        & 0.24 (140.72)             & 0.23 (35.55)   & 0.85 (236.38)       & 1.54 (700.21)          & 14.83 (3924.19)   \\
PGMExplainer & 236.02      & 409.17           & 226.17 & 7,481.89   & 3,457.25      & 6,392.86 \\
OrphicX      & 12.35        & OOM              & 16.15        & OOM          & OOM           & OOM      \\
FastDnX      & 0.11 (66.36) & 0.16 (285.73)    & 0.08 (10.20) & 0.61 (21.55) & Err           & Err      \\
GNNShap      & 104.73       & 179.33           & 33.24        & OOM          & OOM           & OOM      \\
DistShap     & 15.91        & 28.56            & 5.26         & 129.31       & 106.48        & 171.21  \\
\hline
\end{tabular}
}
\end{table*}

\subsection{Explanation Time}
Table~\ref{table:dist-explanation-times} reports the total explanation times for generating explanations for 50 test nodes using various methods. 
In this experiment, DistShap utilizes 128 GPUs for large datasets and 8 GPUs for smaller ones, while all other methods run on a single GPU. 
We acknowledge that this is not a direct, one-to-one comparison. 
However, our goal is to demonstrate that distributed algorithms can significantly reduce the runtime of Shapley value computations, enabling more accurate explanations to be obtained within a practical timeframe.

\begin{figure*}[!t]
    \centering
        \includegraphics[width=1.0\linewidth]{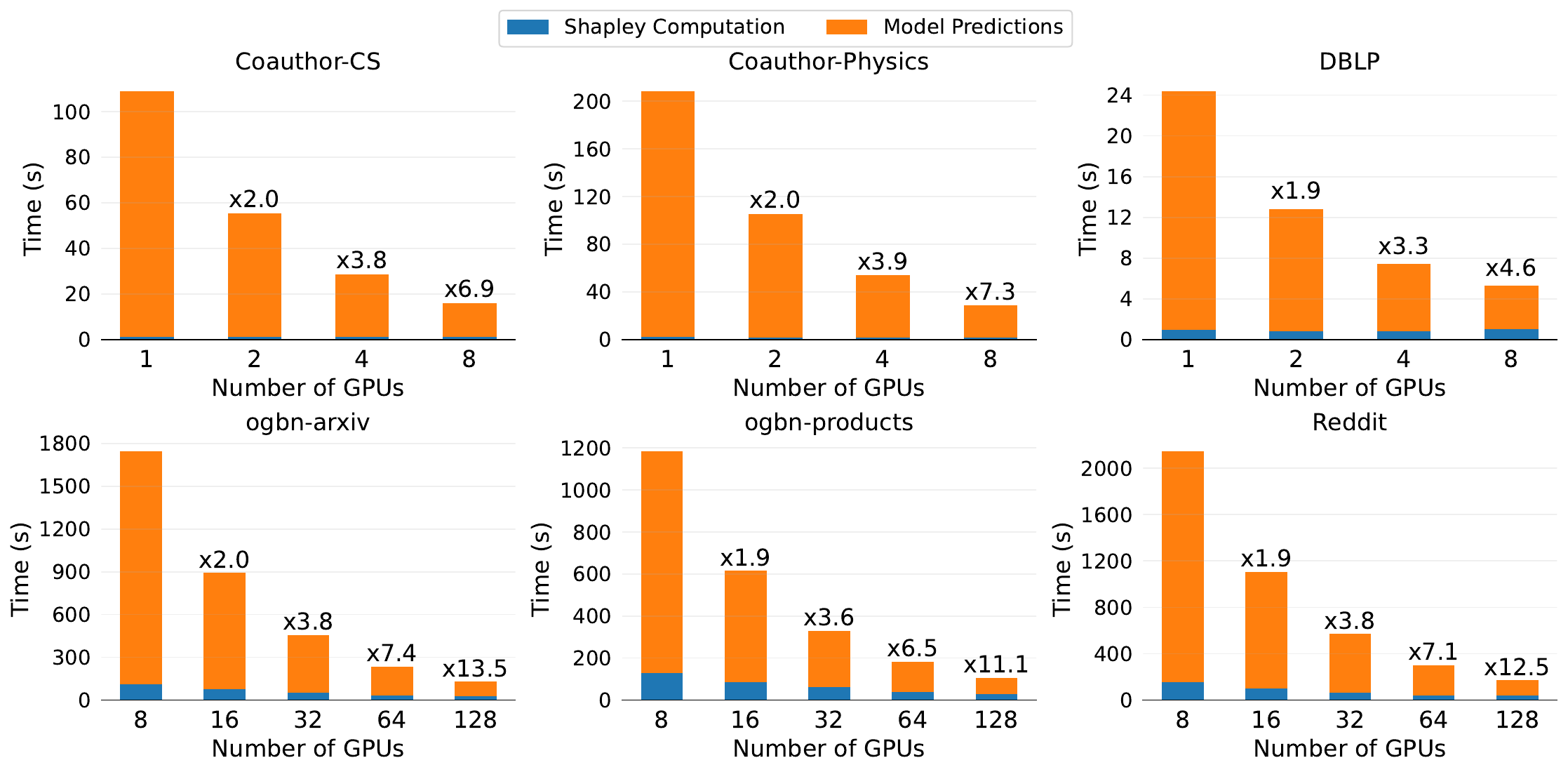}\hfill
        \vspace{-10pt}
        \caption{Scalability of DistShap across different numbers of GPUs. 
        The y-axis denotes the time needed to explain predictions for 50 test nodes. The speedup values above each bar indicate performance improvements relative to the first bar for each respective dataset.
        We use 600,000 samples for large graphs and 60,000 for small ones. }
        \label{fig:dist-scaling}
\end{figure*}

\begin{figure}[!t]
    \centering
        \includegraphics[width=0.9\linewidth]{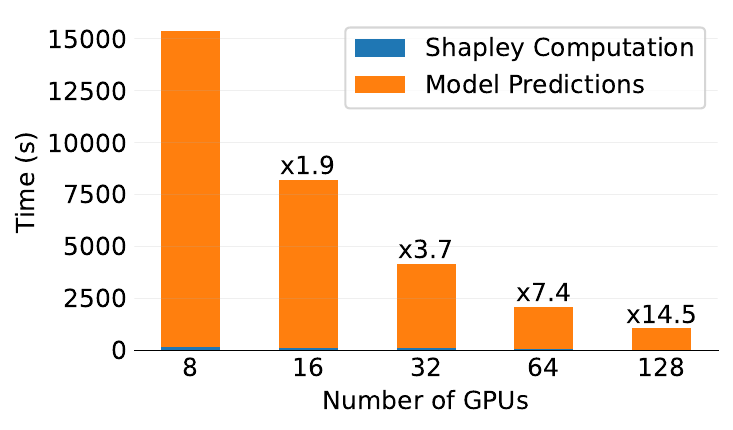}
        \vspace{-5pt}
        \caption{Scalability of DistShap on the Reddit dataset using a two-layer GAT model with 8 attention heads.  Explanation time is measured for 50 nodes and 600K samples per node. \vspace{-5pt} }
        \label{fig:reddit-gat-scalability}
\end{figure}

Table~\ref{table:dist-explanation-times} shows that Saliency is the fastest explanation method, as it requires only a single forward and backward pass through the GNN. 
However, it performs poorly in terms of $Fidelity_{+}$ scores. PGExplainer is also relatively fast, but it incurs additional overhead from training a surrogate model and often yields low fidelity. 
GNNExplainer achieves faster runtime as well, but its fidelity scores remain suboptimal. 
Among all methods, PGMExplainer is one of the slowest, while it achieves reasonable $Fidelity_{-}$ scores, its performance on $Fidelity_{+}$ is comparatively weak.
GNNShap, which is algorithmically similar to DistShap, runs out of memory when applied to large graphs.
Compared to other methods, DistShap delivers accurate explanations within a practical runtime.

\subsection{Strong Scaling of DistShap}
Fig. \ref{fig:dist-scaling} shows the scalability DistShap across different numbers of GPUs. 
For the smaller graphs shown in the top row of Fig.~\ref{fig:dist-scaling}, we report runtime using 1 to 8 GPUs. For the larger graphs in the bottom row, we report runtime from 8 to 128 GPUs.
We present a breakdown of the total runtime into model prediction (including sampling) and Shapley value computation (including solving the least squares problem).

For smaller graphs, DistShap achieves a speedup of $4.6\times$ to $7.3\times$ compared to its runtime on a single GPU.
Here, speedup depends on the single-GPU baseline runtime. For instance, Coauthor-Physics achieves a near-linear speedup of $7.3\times$, benefiting from a relatively high single-GPU runtime of around 200 seconds. In contrast, DBLP shows a $4.6\times$ speedup since its single-GPU runtime is already quite low.
For larger graphs, DistShap achieves a speedup of $11.1\times$ to $13.5\times$ when scaling from 8 to 128 GPUs.
This near-linear speedup is due to the high baseline runtime on 8 GPUs for large datasets.

Across all datasets, we observed that model predictions account for the majority of the runtime. 
This is expected, as an explanation task with 600K samples requires running 600K GCN model predictions on different subgraphs of the computation graph. 
Although we distribute these prediction tasks across GPUs, their sheer volume makes them the primary contributor to the overall runtime. 
As described in the methods section, we assign prediction tasks in a way that avoids inter-GPU communication, enabling model prediction time to scale linearly with the number of GPUs.
In contrast, Shapley value computations using distributed CGLS consume less time overall, but their relative cost increases with the number of GPUs. This is primarily due to the all-reduce communication overhead in the Conjugate Gradient solver, which scales poorly as the number of GPUs increases.

So far, we have focused on explanations with GCN, but DistShap's performance extends to other GNN models as well.
For instance, Fig.~\ref{fig:reddit-gat-scalability} illustrates the scalability of DistShap when explaining 50 test nodes on the Reddit dataset using the GAT model.
We observe even better scalability compared to GCN, as GAT model predictions are more time-consuming than those of GCN.



\begin{figure}[!t]
    \centering
        \includegraphics[width=0.85\linewidth]{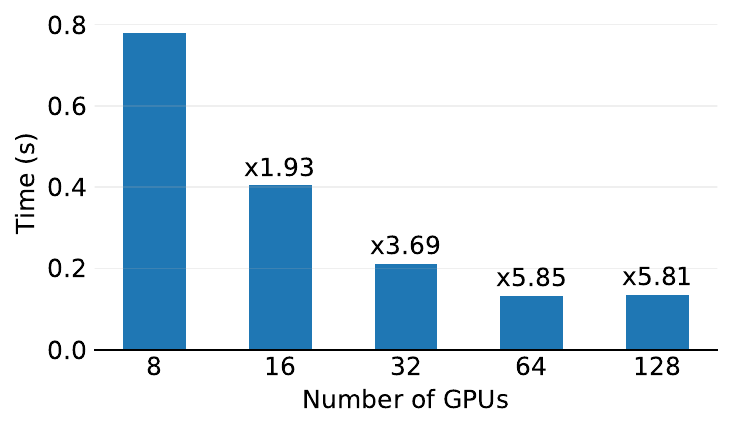}
        \vspace{-10pt}
        \caption{Scalability of sampling on the Reddit dataset. Total sampling time for explaining 50 nodes. \vspace{-5pt} }
        \label{fig:sampling-scalability}
\end{figure}

\subsection{Ablation Studies}\label{sec:ablation-studies}
{\bf Memory overhead of replicated computational graph.}
In our implementation, we replicate the computational graph in every GPU before computing Shapley values.  
For most practical GNNs that use only two or three layers, their computational graphs fit within GPU memory, as shown in the table~\ref{tab:memory}. 
However, deeper GNNs may expand a computational graph that spans a large portion of the input graph, leading to memory overflow. In such cases, memory usage can be managed using neighborhood sampling techniques used in GraphSAGE or GraphSAINT.

\begin{table}[!t]
    \centering
    \caption{Edges and memory requirement to store the largest computation graphs for various datasets. Memory requirement is computed to store a 2-layer computational graph and corresponding features.}
    \begin{tabular}{l r  c }
    \toprule
    Graph & edges in the largest $G_c$ & Memory \\
    \toprule
         Coauthor-CS & 3,299 &  0.04GB \\
Coauthor-Physics & 18,622 & 0.15GB \\
DBLP & 5,014 & 0.01GB \\
ogbn-arxiv & 470,080 & 0.03GB \\
ogbn-products & 2,722,095 & 0.11GB \\
Reddit & 37,787,300 & 1.04GB \\
\bottomrule
    \end{tabular}
    \label{tab:memory}
\end{table}

{\bf Sampling time and scalability.}
Although the sampling step generates 30 million subgraphs for 50 nodes, it remains extremely fast due to our strategy of replicating the computation graph across GPUs.
Fig.~\ref{fig:sampling-scalability} demonstrates that sampling is highly efficient and scales effectively up to 64 GPUs. A minor slowdown is observed at 128 GPUs, which can be attributed to CUDA kernel overhead. 

\begin{figure}[!tb]
    \centering
    \includegraphics[width=1.0\linewidth]{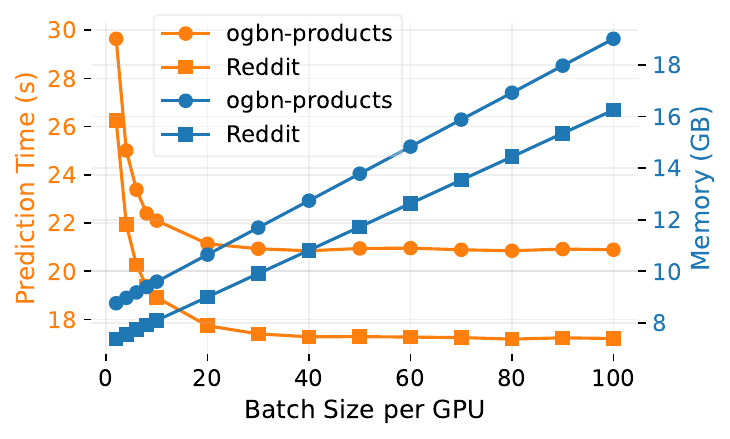}
    \vspace{-15pt}
    \caption{Model prediction times and memory consumption for different batch sizes. Node 18 in Reddit has 92,965 edges and node 236039 in ogbn-products has 97,019 edges in their respective computational graphs.  50,000 samples are used. }
    \label{fig:dist-ablationbatchsize}
\end{figure}


{\bf The impact of batching in model predictions.}
When explaining the prediction of a node, we batch 600K model predictions -- one for each sampled subgraph -- allowing each batch to leverage the parallelism available on each GPU.
Fig.~\ref{fig:dist-ablationbatchsize} shows that model prediction efficiency improves as the batch size increases, with optimal performance reached at a batch size of 50. Beyond this point, GPU utilization becomes saturated, and further increases in batch size yield no additional speedup. This performance gain, however, comes at the expense of higher memory consumption.


{\bf Communication time in Shapley calculations.}
We observed that communication accounts for approximately 2\% of the total runtime of the CGLS solver during Shapley value computation. While communication does not scale as efficiently as computation, its overhead remains minimal -- even when using 128 GPUs -- and thus does not significantly affect the scalability of DistShap.

\begin{figure}[!t]
    \centering
    {\includegraphics[width=0.48\linewidth]{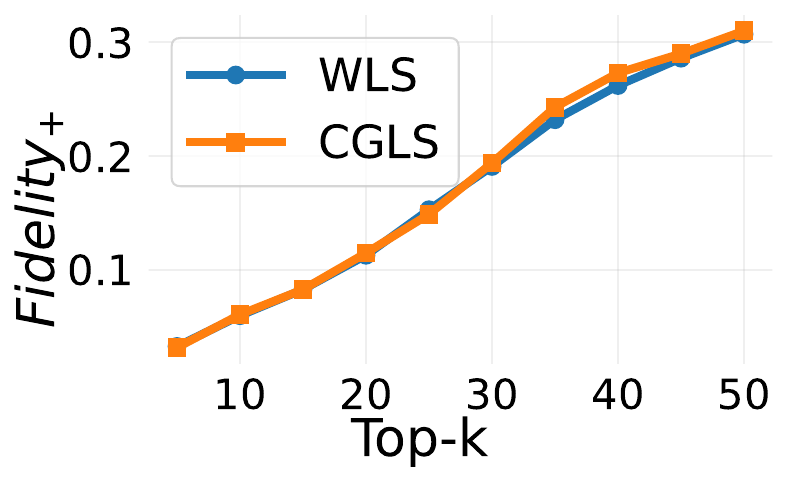}}
  {\includegraphics[width=0.48\linewidth]{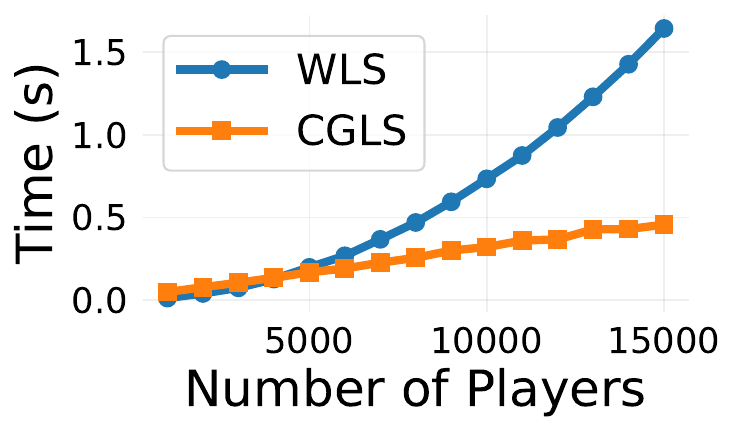}}
        \vspace{-5px}
        \caption{(left) WLS and CGLS $Fidelity_{+}$ score comparison on Coauthor-CS dataset. (right) WLS and CGLS computation times for 60,000 samples with varying players.}
        \label{fig:dist-ablcglswls}
\end{figure}

{\bf Accuracy and runtime of  WLS vs CGLS.}
We experimented with PyTorch's native WLS solver but found it to be unscalable as the number of players increased.
As shown in Fig.~\ref{fig:dist-ablcglswls}, both WLS and our CGLS method yield similar fidelity, but CGLS runs significantly faster as the number of players grows.

\section{Conclusions}
We introduce DistShap, a scalable GNN explanation method using distributed Shapley value computation. DistShap improves over prior methods in two key ways: (1) it achieves high-fidelity explanations for large computational graphs by distributing millions of non-uniform samples across multiple GPUs, and (2) it contributes to HPC with efficient subgraph sampling, batched model prediction, and a scalable Conjugate Gradients Least Squares (CGLS) solver. 
All of these components scale nearly linearly to 128 GPUs and can be used by any other applications.

The distributed version of GNNShap marks significant progress, but it still leaves several areas open for improvement.
In our experiments, we used a two-layer GNN as it delivers state-of-the-art performance for these datasets. However, deeper GNNs are often beneficial for heterophilic graphs, where neighbors have different node labels. 
In such cases, the computational graph may encompass a large portion of the entire graph. With deeper GNNs, our approach of replicating the computational graph on GPUs could lead to memory bottlenecks.
Our future work will focus on explaining deeper GNNs and expanding the application of DistShap to other deep learning models.

\section{Acknowledgments}
This research was funded in part by DOE grants DE-SC0022098 and DE-SC0023349, and by NSF grants CCF-2316233 and OAC-2339607.





\bibliographystyle{ACM-Reference-Format}
\bibliography{references}

\appendix




\end{document}